\documentclass[conference]{IEEEtran}
\IEEEoverridecommandlockouts
\usepackage{cite}
\usepackage{amsmath,amssymb,amsfonts}
\usepackage{algorithmic}
\usepackage{graphicx}
\usepackage{textcomp}
\usepackage{xcolor}
\usepackage{todonotes}
\usepackage{threeparttable}
\def\BibTeX{{\rm B\kern-.05em{\sc i\kern-.025em b}\kern-.08em
    T\kern-.1667em\lower.7ex\hbox{E}\kern-.125emX}}
\begin{document}

\title{Exploring grid topology reconfiguration using a simple deep reinforcement learning approach
}

\author{\IEEEauthorblockN{Medha Subramanian}
\IEEEauthorblockA{\textit{Dept.~of Electrical Sustainable Energy} \\
\textit{Delft University of Technology}\\
Delft, The Netherlands\\
medha.subramanian@smartwires.com${^\$}$ \thanks{\textsuperscript{\$}
The research was carried out as an intern at TenneT for an MSc thesis project. 
MS is currently employed at Smart Wires Inc.}}
\and
\IEEEauthorblockN{Jan Viebahn}
\IEEEauthorblockA{\textit{Digital and Process Excellence} \\
\textit{TenneT TSO B.V.}\\
Arnhem, The Netherlands \\
jan.viebahn@tennet.eu}
\and
\IEEEauthorblockN{Simon H. Tindemans}
\IEEEauthorblockA{\textit{Dept.~of Electrical Sustainable Energy} \\
\textit{Delft University of Technology}\\
Delft, The Netherlands \\
s.h.tindemans@tudelft.nl}
\and[\hfill\mbox{}\par\mbox{}\hfill]
\IEEEauthorblockN{Benjamin Donnot}
\IEEEauthorblockA{
\textit{RTE}\\
Paris, France \\
benjamin.donnot@rte-france.com}
\and
\IEEEauthorblockN{Antoine Marot}
\IEEEauthorblockA{
\textit{RTE}\\
Paris, France \\
antoine.marot@rte-france.com}
}

\IEEEpubid{\parbox{\columnwidth}{\copyright 2021 IEEE. Personal use of this material is permitted. Permission from IEEE must be obtained for all other uses, in any current or future media, including reprinting/republishing this material for advertising or promotional purposes, creating new collective works, for resale or redistribution to servers or lists, or reuse of any copyrighted component of this work in other works.}\hspace{\columnsep}\makebox[\columnwidth]{ }}

\maketitle

\IEEEpubidadjcol

\begin{abstract}
System operators are faced with increasingly volatile operating conditions. In order to manage system reliability in a cost-effective manner, control room operators are turning to computerised decision support tools based on AI and machine learning. Specifically, Reinforcement Learning (RL) is a promising technique to train agents that suggest grid control actions to operators. In this paper, a simple baseline approach is presented using RL to represent an artificial control room operator that can operate a IEEE 14-bus test case for a duration of 1 week. This agent takes topological switching actions to control power flows on the grid, and is trained on only a single well-chosen scenario. The behaviour of this agent is tested on different time-series of generation and demand, demonstrating its ability to operate the grid successfully in 965 out of 1000 scenarios. The type and variability of topologies suggested by the agent are analysed across the test scenarios, demonstrating efficient and diverse agent behaviour.
\end{abstract}


\section{Introduction}

Operators of electricity transmission and distribution networks are responsible for the safe and reliable operation of these networks. This task grows progressively more challenging due to the increasing presence of renewable generation and power-electronics-based resources on the grid. Both trends conspire to make power flows in the network more variable, less predictable and sensitive to disturbances. Network control rooms are staffed by operators who rely on their experience to anticipate and resolve undesirable system behaviour. Faced with an increasingly volatile network, continuing this mode of operation is likely to reduce system security, or greatly increase costs of operation. 

In recognition of this trend, there is an increasing reliance on ICT and smart grid technology in the control room \cite{stevens2015situation}. In particular, researchers and system operators have proposed machine learning to anticipate risks, notably in the context of online dynamic security assessment (transient stability analysis) 
\cite{Wehenkel1991, konstantelos2019}. However, even when the performance of such anticipatory systems is adequate, additional complexities must be tackled when they are used as the basis for operator decisions \cite{Cremer2019}.

In recent years, Reinforcement Learning (RL) has emerged as a powerful approach to automated decision making. Through iterative experimentation, RL arrives at policies that aim to maximise a numerical reward signal. The basic RL framework is applicable to a tremendous range of settings including the control of electricity grids \cite{kelly2020reinforcement}. One important initiative in the context of RL for grid operation are the \emph{Learning To Run a Power Network} (L2RPN) challenges organised by RTE \cite{l2rpn}, the French Transmission System Operator. This competition was conducted with the primary goal of introducing and recognising the potential of AI and ML based tools to support control rooms and assist in making optimal decisions. The first edition of this competition was held in 2019 \cite{marot2019learning}. The L2RPN competition focuses on learning a subset of actions that can be taken by grid operators, namely topological changes. These can often be implemented at no or low cost, and provide significant flexibility to avoid violation of operational constraints by manipulating power flows in the grid.

It is tempting to extrapolate these efforts and envision RL agents that control the grid autonomously. However, for the control of critical infrastructure it is both undesirable to place responsibility in the `hands' of an algorithm and -- for the foreseeable future -- it is infeasible to achieve the required degree of accuracy and dependability. Instead, the goal is to develop RL agents that can act \emph{in conjunction with} human network operators in the form of decision support tools. In particular, such agents may identify and suggest control actions that human operators and traditional solution techniques are unaware of or unaccustomed to \cite{prostejovsky2019future,kelly2020reinforcement}.

This paper describes a lightweight approach to train agents of the latter type, specifically one that proposes topology control actions to a system operator. Such an agent could serve as a baseline for other studies with more elaborate methods. The main contributions of this paper are:
\begin{itemize}
\item Development of a simple, yet mostly successful, cross-entropy agent that is trained on only a single well-chosen scenario.
\item In-depth analysis of the performance and sequential decisions made by this agent.
\end{itemize}

\section{Power System Framework}\label{power_system}

\subsection{Power grid model}
\begin{figure}
    \centering
    \includegraphics[scale=0.3]{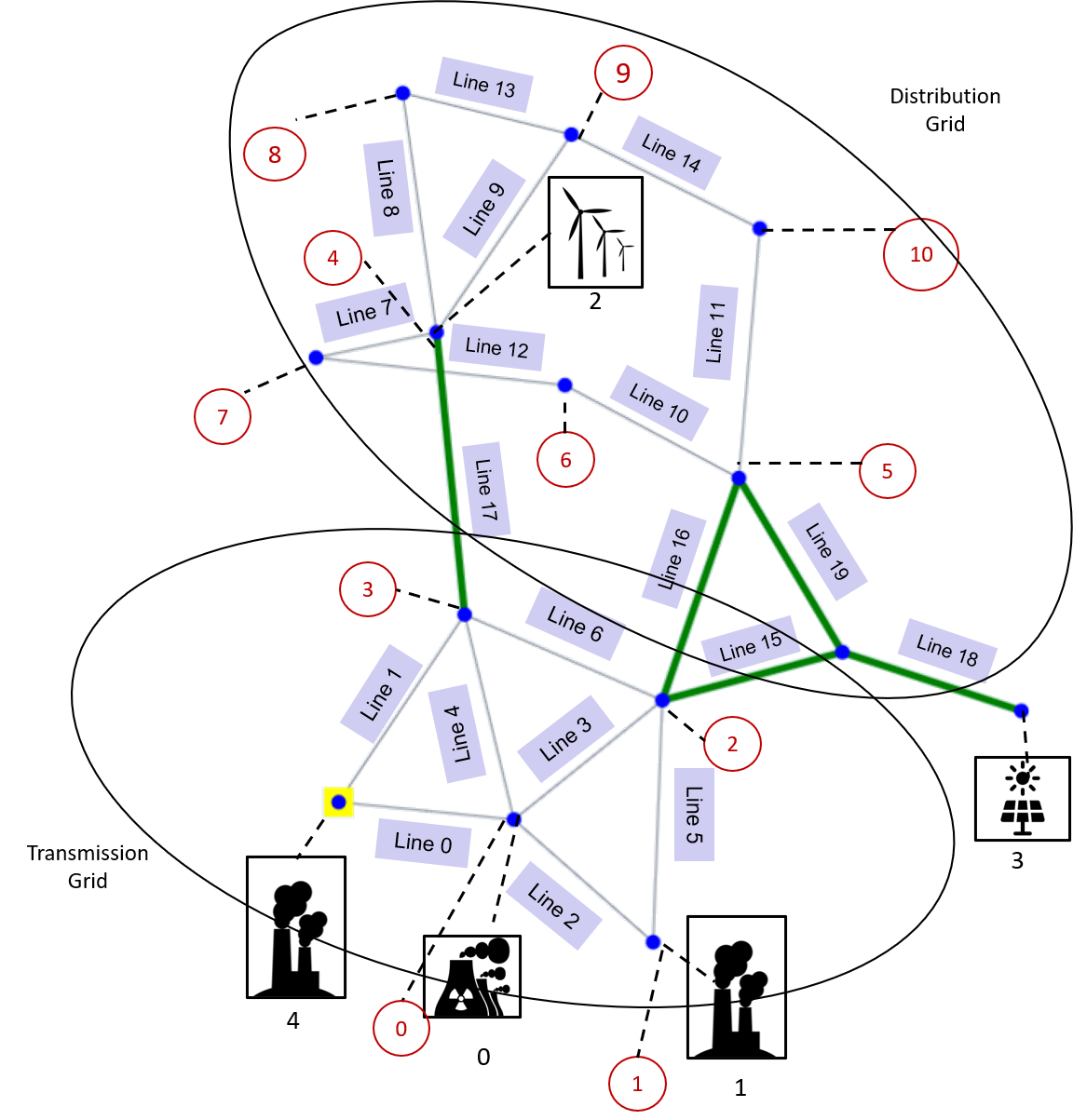}
    \caption{The adapted IEEE 14 bus network including substations (blue dots),
    loads (red circles), and generators (panels). Green lines represent transformers.}
    \label{fig:14_levels}
\end{figure}

The power grid model considered in this study is a slightly adapted version of the IEEE 14-bus network, as it was created for the L2RPN challenge 2019 \cite{l2rpn}.
Figure \ref{fig:14_levels} sketches the main elements of the grid: 
20 lines, 11 loads and 5 generators.
Generation includes a wind power plant and a solar in-feed next to
a nuclear generator and two thermal generators to represent the current energy mix.
We modified part of the thermal current limits of the lines, as shown in Table \ref{tab:thermal_lts},
to make the difference between the transmission (lower) and distribution (upper) sections of the grid (indicated by the two circles in Fig.~\ref{fig:14_levels}) more pronounced.
The transformers to step down the voltage from the transmission side to the distribution side are modelled as lines and are represented in green in Fig.~\ref{fig:14_levels}.

\begin{table}
\centering
\caption{Line thermal limits}
\label{tab:thermal_lts}
\begin{tabular}{|l|l|l|l|}
\hline
\textbf{Line} & \textbf{\begin{tabular}[c]{@{}l@{}}Thermal\\  limit (A)\end{tabular}} & \textbf{Line} & \textbf{\begin{tabular}[c]{@{}l@{}}Thermal \\ limit (A)\end{tabular}} \\ \hline
0 & 1000 & 10 & 380 \\ \hline
1 & 1000 & 11 & 380 \\ \hline
2 & 1000 & 12 & 760 \\ \hline
3 & 1000 & 13 & 760 \\ \hline
4 & 1000 & 14 & 380 \\ \hline
5 & 1000 & 15 & 760 \\ \hline
6 & 1000 & 16 & 380 \\ \hline
7 & 760 & 17 & 380 \\ \hline
8 & 450 & 18 & 2000 \\ \hline
9 & 760 & 19 & 2000 \\ \hline
\end{tabular}
\end{table}

The power grid model is available within the Python module \emph{Grid2Op} \cite{grid2op}
that provides an environment for development, training or evaluation of `agents' or `controllers' that act on a power grid in different ways.
It uses \emph{Pandapower} \cite{thurner2018pandapower} as a backend for power flow computations
and the package is compatible with Open-AI gym \cite{openai_gym_nodate}.
The module also comes with datasets representing realistic time-series of operating conditions.
The dataset for the IEEE 14-bus model contains 1000 scenarios with data for 28 continuous days
in a 5-minute interval.
Each scenario includes pre-determined load variations and generation schedules
which are representative of distributions of the French grid \cite{marot2019learning}.
In this study only the first week of each scenario will be used.

\subsection{Objective and constraints}\label{objective_constraints}

The main objective is to create an agent that is able to operate the power grid
successfully for as many scenarios as possible, using only 
topology adjustment actions (described in detail in section \ref{action_space}).
In doing so, the agent must respect a number of  operational constraints \cite{grid2op}.
\emph{Hard constraints}, which trigger an immediate ``game over'' condition if violated, are:
\begin{itemize}
    \item[(a)] system demand must be fully served; 
    \item[(b)] no generator may be disconnected;
    \item[(c)] no electrical islands are formed as a result of topology control;
    \item[(d)] AC power flow must converge at all times.
\end{itemize} 
In contrast, \emph{soft constraints} have less severe consequences: Transmission lines with a current exceeding 
150\% of their rated capacity are tripped immediately, and can be
recovered after 50 minutes (10 time steps).
When lines are overloaded by a smaller amount, the agent has 10 minutes (2 time steps) to mitigate this. If lines remain overloaded after this time, they are disconnected and will be  reconnected after 50
minutes. In addition, substations are subject to a practical `complexity' constraint that  only
one substation can be modified per timestep
and a `cooldown time' (15 minutes) needs to be respected
before a switched node can be reused for action. Both soft and hard
constraints make the problem more practical and close to
real-world grid operation. 

We note that, in this study, only quasi-steady state load-flows are used. Other aspects, such as dynamic performance, N-1 contingency analysis, or voltage performance are not considered.

\section{Reinforcement Learning Approach}\label{RLapproach}
The intention of this study is to apply a rather simple
RL approach which is easy to understand and can serve as a baseline for more advanced algorithms.
Hence, we consider a severely reduced action space (\ref{action_space}),
a simple neural network architecture (\ref{obs_reward_nn}),
a simple learning algorithm (\ref{cem}), 
and small training data (\ref{training}).
We note that the formulation of the control problem considered here as a Markov Decision Process (MDP) can be found e.g. in \cite{marot2019learning,Lan2019}.

\subsection{Action Space}\label{action_space}
In this study we consider only one type of action, namely \emph{bus-bar splitting}.
In reality a series of actions is necessary in order to carry out bus splitting and to obtain an intended substation configuration. 
However, for the RL agent we focus only on the final substation configuration. 
That is, the agent directly selects the intended substation configuration for a given network node.
In the following, we briefly describe the action space constraints 
and we provide a formula to compute the number of permissible configurations per substation.

The overall number of configurations of a specific substation is dependent on the number of elements connected to the substation. However, not all configurations are consistent with power grid operations. In this study the following constraints are incorporated:
\begin{enumerate}
    \item \textbf{Minimum Element Constraint:} A minimum of two elements (or 0 elements) must be connected to each bus-bar.  
    \item \textbf{One-Line Constraint:} At least one of the elements connected to a bus-bar must be a line. That is, it is not permitted to connect only non-line elements (generators and loads) to a bus-bar (as implied by constraint (c) in section \ref{objective_constraints}).
\end{enumerate}

We note that these constraints are \textit{local} since they only refer to the configuration of each individual substation. \textit{Global} constraints that refer to the overall grid topology induced by the configuration of 2 or more substations are not considered by the agent.

The total number of configurations $\tau$ of a substation satisfying the two local constraints can be calculated as follows:
\begin{equation}
    \tau= \alpha(n) - \beta(n) - \gamma(n^\prime)\ ,
    \label{eq:configs_calc}
\end{equation}
with $\alpha(n)=2^{n-1}$, $\beta(n)=n - \delta_{n,2}$, and
$\gamma(n^\prime) = 2^{n'} - 1 - n'$,
where $n$ is the total number of elements connected to the substation,
$n^\prime$ is the number of non-line elements connected to the substation and $\delta_{n,2}$ is the Kronecker delta that equals 1 when $n=2$ and 0 otherwise.

The reasoning is as follows:
The term $\alpha$ gives the number of all possible configurations of a substation with two bus-bars and $n$ elements without taking the local constraints into account. An additional factor of $\frac12$ corrects for double-counting of symmetrical configurations that arises because the two bus-bars are indistinguishable.
The two terms $\beta$ and $\gamma$ reduce the number of configurations according to the two constraints.
The $\beta$ term counts the number of substation configurations in which only 1 element (or the symmetrical version, that is, $n-1$ elements) is connected to a bus-bar (i.e. constraint 1).
Finally, the $\gamma$ term implements constraint 2. It subtracts all states where the lines are connected to a single bus-bar and the non-line elements are distributed in any way other than to the same bus-bar ($2^{n'}-1$). The term $-n'$ is included to avoid double-counting with a single element (already accounted for in $\beta$). 

With \eqref{eq:configs_calc} the number of valid configurations for each substation can be calculated as seen in Table \ref{tab:configs}. 
In the power system model (section \ref{power_system}) at a single time-step only one substation can be acted upon. Thus, at each time-step, only one of these 112 substation configurations can be chosen as an action.
We note that each number in table \ref{tab:configs} also includes one do-nothing action, namely,
when the substation configuration that is already in place is chosen.
Hence, the total number of unitary actions excluding do-nothing actions is $112-14=98$.
Note that the combination of substation configurations results in over 23 million possible grid topologies (some of which may result in electrical islands).

\begin{table}
\centering
\caption{Number of possible configurations for each substation}
\label{tab:configs}
\begin{tabular}{|l|l|l|l|}
\hline
\textbf{\begin{tabular}[c]{@{}l@{}}Substation \\ number \end{tabular}} & \textbf{\begin{tabular}[c]{@{}l@{}}Number \\ of  elements \end{tabular}} & \textbf{\begin{tabular}[c]{@{}l@{}}Number \\ of  configurations \end{tabular}}\\ \hline
0 & 3 & 1       \\ \hline
1 & 6 & 25   \\ \hline
2 & 4 & 3 \\ \hline
3 & 6 & 26  \\ \hline
4 & 5 & 11  \\ \hline
5 & 6 & 25  \\ \hline
6 & 3 & 1   \\ \hline
7 & 2 & 1  \\ \hline
8 & 5 & 11 \\ \hline
9 & 3 & 1 \\ \hline
10 & 3 & 1 \\ \hline
11 & 3 & 1 \\ \hline
12 & 4 & 4  \\ \hline
13 & 3 & 1 \\ \hline
 &\textbf{\begin{tabular}[c]{@{}l@{}}Total substation \\ configurations\end{tabular}} &  \textbf{112}                               \\ \hline
\end{tabular}%
\end{table}

\subsection{Observations, neural network, reward}\label{obs_reward_nn}
Before we can consider the actual learning algorithm in the next section
we need to specify the observations provided to the agent at each timestep,
the neural network architecture used to compute the action probabilities, and the reward function.

The agent receives a partial observation of the power system state.
We did not apply any reduction technique to the state space, opting to include almost all available observations at the current time step, but no observations from previous time steps, i.e. no memory.
The agent observes the following 324 features:
generators' voltages (V), active powers (P) and reactive powers (Q) $(5 \times 3 = 15)$,
loads' P, Q, V $(11 \times 3 = 33)$,
lines' P, Q, V, I for both origins and extremities $(40\times 4 = 160)$,
lines' loading $(20)$,
topology vector (bus-bar index for each element: $5+11+40 = 56$),
line status $(20)$, and
number of time steps a line is overloaded $(20)$.

In order to model the policy of the agent we use a simple feed-forward neural network (NN) with two hidden layers.
The input layer of the NN has the size of the state space (i.e. 324). Each hidden layer consists of 300 neurons. The size of the output layer corresponds to the size of the action space (i.e. 112).

Finally, we use the same reward function as defined in \cite{marot2019learning,Lan2019}. That is, the immediate reward at each time step reflects the remaining available transfer capabilities.
It combines the loading of all lines and it increases (decreases) if the loading of the network decreases (increases). Note that this indirectly favours successful operation of the grid, because (i) overloads are discouraged and (ii) a ``game over'' case stops the accumulation of reward.
For more details we refer to \cite{marot2019learning,Lan2019}.

\subsection{Cross-Entropy Method}\label{cem}
The cross-entropy method (CEM) is a Monte Carlo technique based on importance sampling; it belongs to the class of model-free, policy-based, on-policy RL methods \cite{rubinstein1997optimization,lapan2018deep}.
Its main strength is simplicity: it is intuitive and easy to implement. 
In essence, the method samples a number of episodes
and subsequently selects the best ones for training. 
Note that an \emph{episode} is a sequence of states, actions, and rewards from an initial state to a terminal state in which the agent either succeeded or failed to operate the grid for the duration of a specific scenario (see \ref{objective_constraints}).

\begin{figure}
    \centering
    \includegraphics[scale = 0.08]{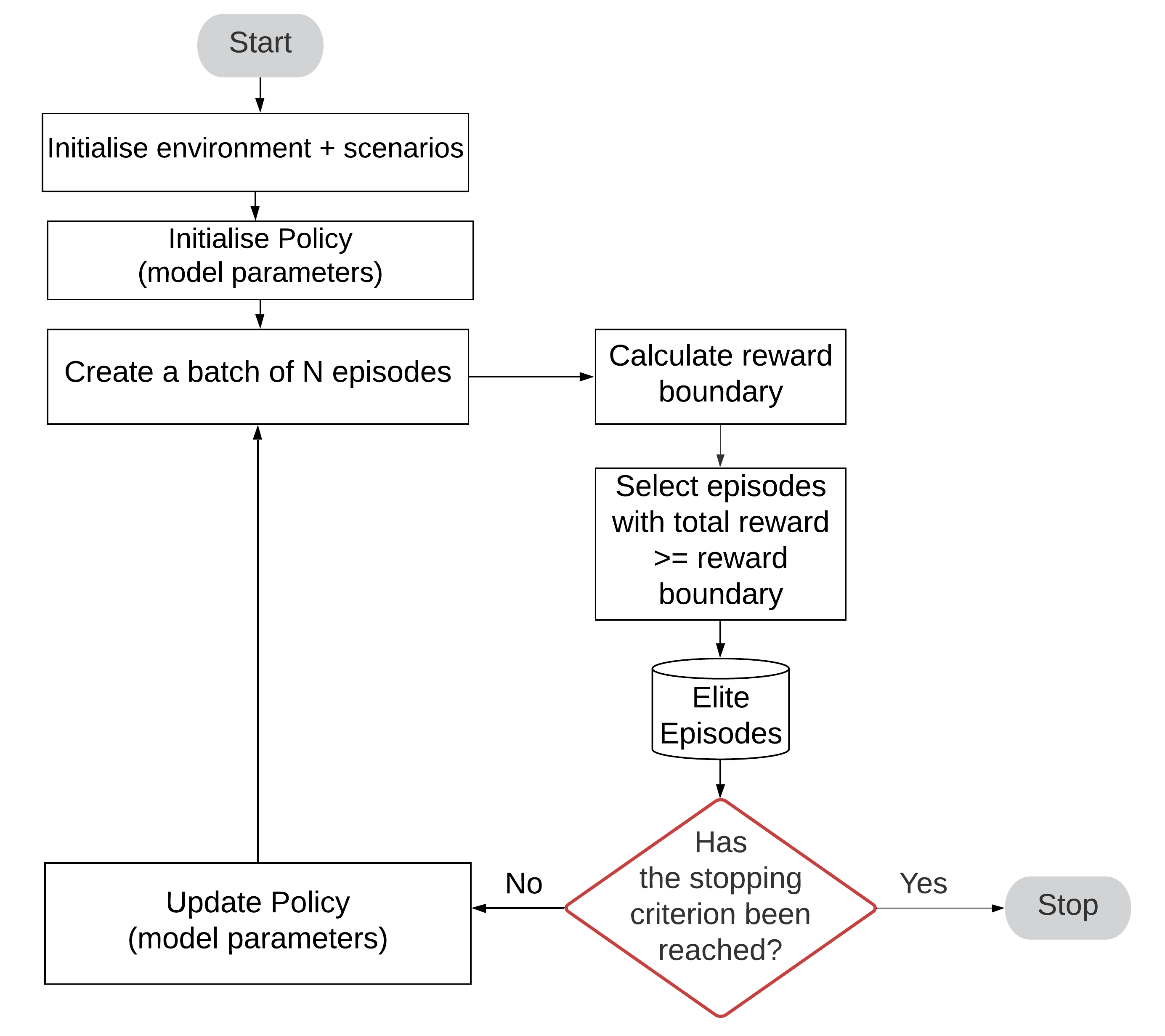}
    \caption{Implementation logic of the Cross-Entropy Method.}
    \label{fig:CEM_flow}
\end{figure}

More precisely, Figure \ref{fig:CEM_flow} shows a flowchart illustrating the logic of the CEM.
A batch is created by generating $N$ episodes using the current policy.
For each episode, the total reward is computed and a set of \emph{elite episodes} is identified: all episodes with a total reward that exceeds a predefined reward boundary (usually some percentile of all batch rewards).
If the stopping criterion has not been satisfied, the policy is updated (i.e. the NN is trained via supervised learning) using the elite episodes and cross-entropy loss.
Subsequently, the next batch is created using the updated policy and the loop starts again.

The training is manually stopped when the reward boundary has (visually) saturated
and the difference between reward boundary and mean reward is less than 2\%
(see e.g. Fig. \ref{fig:training} in the next section).
The hyperparameters used for training are $N=20$ as batch size,
the 75th percentile as reward boundary (i.e. the elite episodes are the 5 best episodes of a batch), and a learning rate of 1e-4.
For details of the NN we refer to section \ref{obs_reward_nn}.

Finally, we adapt the the vanilla CEM in two ways.
A simple but crucial modification is that we include
an \emph{activity threshold} ($AT=0.95$) which is inspired by the warning flag used in \cite{Lan2019}. 
That is, the agent only executes actions if the current highest line loading in the power grid exceeds $0.95$,
otherwise the grid topology remains unchanged.
Applying an activity threshold significantly improves the learning behaviour
since learning of appropriate behaviour in low-loading situations can be omitted. 
The second and less important modification is that we make batch size
and reward boundary slightly dependent on the results of the current batch and the previous batch.
Specifically, we assure by continued sampling that either the total reward of the 5\textsuperscript{th} best episode (i.e. the new reward boundary) exceeds the previous reward boundary
or the 5 best episodes all complete successfully.
The effect of this modification is 
to make the learning curves (see e.g. Fig. \ref{fig:training} in the next section) a bit smoother.

\section{Results}
\subsection{Training of the agent}\label{training}
In this study we specify the training data not in a statistical way
but guided by expert knowledge, which simplifies the training procedure and drastically reduces its computational burden.
To select the training scenario we consider the \emph{base topology}
in which for each substation all elements are connected to busbar 1 (this is also the default topology in \emph{Grid2Op} as well as a topology preferred by real operators).
We compute the load flow for each scenario
with the grid topology being fixed to the base topology (i.e. a do-nothing agent) and without line tripping due to overloading (see section \ref{objective_constraints}).

\begin{figure}
    \centering
    \includegraphics[scale = 0.55]{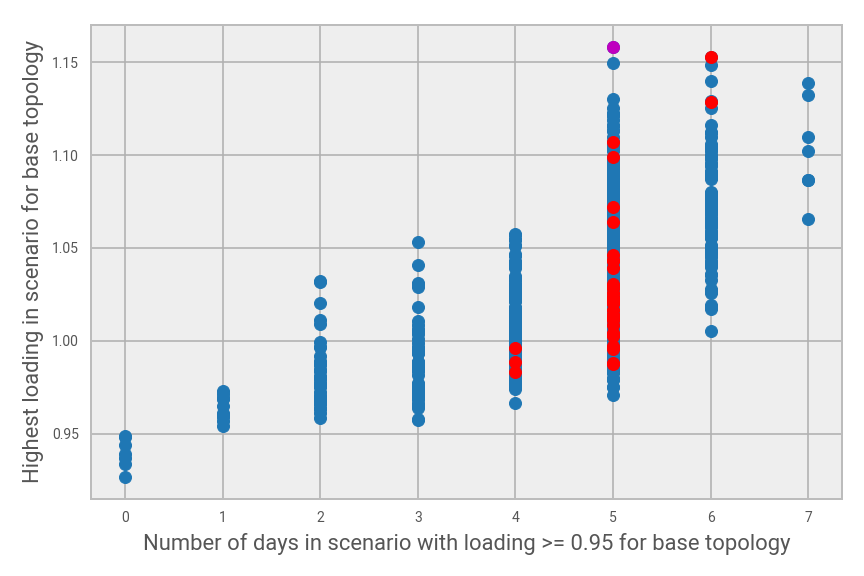}
    \caption{For each of the 1000 1-week-long scenarios, the number of days
    with loading $\geq 0.95$ (horizontal) and the highest loading (vertical) are shown. The magenta dot is related to the scenario used for training. The 35 red dots are related to scenarios which the agent cannot operate successfully.}
    \label{fig:overloading_days_vs_max}
\end{figure}

Figure \ref{fig:overloading_days_vs_max} shows for each of the 1000 1-week-long scenarios the number of days with loading $\geq 0.95$ and the highest loading for the entire scenario.
There are a number of scenarios where the highest loading remains smaller than 1 for the entire duration. There is even a small number of scenarios where the highest loading stays below 0.95 (i.e. the blue dots related to 0 days).
Consequently, when line tripping due to overloading (see section \ref{objective_constraints}) is applied then simply sticking to the base topology leads to episodes of successful operation for about 30\% of the scenarios (namely for 297).

In general, Fig. \ref{fig:overloading_days_vs_max} depicts a correlation between the number of days in a scenario with loading above 0.95 and the maximum load level.
Further analysis shows that for each scenario the overloading (i.e. loading larger than 1) is exclusively related to line 9 (see Fig.~\ref{fig:14_levels}).
That is, System Operators would face the question whether new investments are
necessary or other measures like topological reconfiguration are sufficient
to overcome line 9 becoming a bottleneck.

In this study we train an agent using our RL approach (see section \ref{RLapproach})
\emph{on a single scenario}, namely,
on the scenario that appears to be most challenging in terms
of maximum overloading as indicated by the magenta dot in Fig.~\ref{fig:overloading_days_vs_max}.

\begin{figure}
    \centering
    \includegraphics[scale = 0.55]{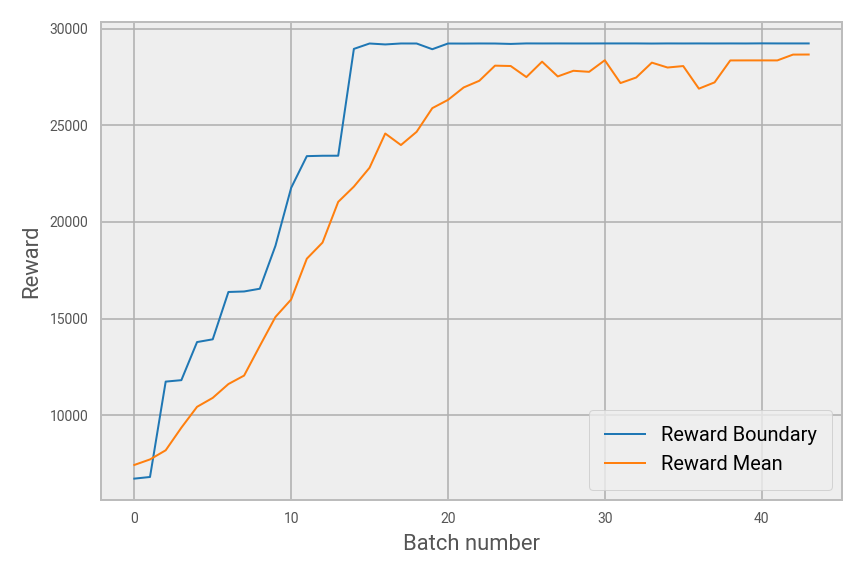}
    \caption{Training progress of the agent monitored by considering
both the mean reward and the reward boundary of the reward distribution of each batch.}
    \label{fig:training}
\end{figure}

Figure~\ref{fig:training} shows the training progress of the agent monitored by considering
both the mean reward and the reward boundary of the reward distribution of each batch.
Both measures increase monotonically for the first 10-15 batches, indicating
improved performance of the agent after each training step.
After around 14 batches the reward boundary reaches a plateau which reflects
that for all elite episodes the agent successfully operates the grid.
After around 23 batches the mean reward also reaches a plateau at a slightly
lower reward value than the reward boundary.
This indicates that for almost all episodes of a batch the agent successfully operates the grid.
The slightly lower value of the mean reward as well as its small oscillations
are due to the remaining exploration probabilities which lead to a small number of failure episodes.

\subsection{Evaluation of the agent}
For evaluating the performance of the agent we recall the setting in which it is to be used, namely not as an independent decision maker, but as a decision support tool that suggests any number of possible actions to a system operator, who can verify these actions before implementing them. In this context, the key performance criterion is whether a desirable action is included in the episodes associated with a given scenario.

To evaluate the agent we generated more than 33000 episodes 
where the injections of each episode were randomly chosen from one of the 1000 scenarios.
This way we obtained at least (at most) 14 (40) episodes per scenario
(episodes with the same scenario can differ due to the probabilistic nature of the agent).
We found that the agent was able to create episodes of successful operation
for 965 scenarios of the 1000 scenarios (including the scenario used for training),
that is, for 96.5$\%$ of the 1000 scenarios.
The 35 scenarios that the agent could never operate successfully
are indicated by the red dots in Fig.~\ref{fig:overloading_days_vs_max}.
We note that the choice of the training scenario is crucial
since choosing other training scenarios can lead to drastically smaller
success rates (results not shown).

In this study we did not investigate how the performance of the agent could be further improved, but we note that it is likely that the remaining
35 scenarios can also be operated successfully using busbar splitting.
This is obvious for the scenarios where the highest loading remains smaller than 1 in the default topology
(see Fig.~\ref{fig:overloading_days_vs_max}).
Moreover, we trained another agent using our RL approach (see section \ref{RLapproach})
using the scenario related to
the red dot in Fig.~\ref{fig:overloading_days_vs_max} with the highest overloading 
as training scenario
and then we also found episodes of successful operation for the respective scenario.

Next, we take a more detailed look at the actions taken by the agent in different episodes of successful operation.
For that we consider for each of the 965 scenarios the episode with the highest reward.
Table \ref{tab:topologies} summarises the different topologies created by the agent as well as the different sequences of their encounter.
Note that the 7 episodes that entirely remain in the base topology are related to the
scenarios for which the highest loading is smaller than 0.95 (see Fig.~\ref{fig:overloading_days_vs_max}).

The agent's actions mostly (see Tab.~\ref{tab:topologies} and also Fig.~\ref{fig:14_levels}) take place at either substation 3 (for $\mathcal{T}2$, $\mathcal{T}5$, and $\mathcal{T}7$) or at substation 8 (for $\mathcal{T}1$) or at both (for all other topologies). Only for 10 episodes substation 1 is additionally involved (for $\mathcal{T}4$).
For about one third of the episodes (namely 342) only one topology change is performed, namely,
either $\mathcal{B}\rightarrow\mathcal{T}1$ or $\mathcal{B}\rightarrow\mathcal{T}2$.
We note that the 37 episodes where $\mathcal{B}\rightarrow\mathcal{T}2$ happens are all related to scenarios for which the highest loading in the base topology is smaller than 1 (see Fig.~\ref{fig:overloading_days_vs_max}).
On the other hand, the episodes where $\mathcal{B}\rightarrow\mathcal{T}1$ happens are related to scenarios for which the highest loading in the base topology is larger than 1 (not shown).

For the majority of episodes (namely 601) two topology changes are performed leading to $\mathcal{T}3$
which combines both $\mathcal{T}1$ and $\mathcal{T}2$.
The remaining small number of episodes also involve
$\mathcal{T}1$ and/or $\mathcal{T}2$ in some form.
In particular, we note that the two episodes with four topological changes actually only involve four or less topologies,
and hence, represent episodes in which topologies are revisited.
Revisitation of topologies only happens for three of the 965 episodes.

Finally, we quantify the extent to which different topological states are present during an episode. 
One way to measure this is by considering the \emph{topological entropy},
which is computed by turning the time spent in a given topological state (divided by the episode length)
into an effective probability of encountering this topology in the respective episode.
That is, it effectively measures the topological uncertainty of an episode
and it is also shown in Table \ref{tab:topologies}.
For example, if the topology is fixed for the entire episode then
the topological uncertainty is zero.
On the other hand, the more topologies are encountered with similar probabilities
the higher the topological entropy.
Consequently, the largest entropy is found for episodes with three topological changes
($\mathcal{B}\rightarrow\mathcal{T}2\rightarrow\mathcal{T}3\rightarrow\mathcal{T}4$).

The main result here is the large variation of topological entropy
across different episodes, showing that this simple agent is able to exhibit a diverse range of behaviour.
For the episodes with a single topological change $\mathcal{B}\rightarrow\mathcal{T}1$,
the topological entropy can be very small (0.0043, indicating that the topological change
happened very early or very late in the episode)
or relatively large (indicating that the topological change happened near the middle of the episode). The same holds for the episodes with the two topological changes $\mathcal{B}\rightarrow\mathcal{T}2\rightarrow\mathcal{T}3$.
On the other hand,
for the single (double) topological changes $\mathcal{B}\rightarrow\mathcal{T}2$ ($\mathcal{B}\rightarrow\mathcal{T}1\rightarrow\mathcal{T}3$) the entropy range is much more confined.
Here, we only considered the minimum and maximum observed entropy values. Future work could investigate the distribution and scenario-dependence of the topological uncertainty.

\begin{table}
\centering

\caption[alternate caption]{Topologies encountered in highest reward episodes:

\begin{itemize}
    \item \textit{$\mathcal{B}$: base topology}
    \item \textit{$\mathcal{T}1$: load 5 and line 19 at busbar 2 (at substation 8)}
    \item \textit{$\mathcal{T}2$: load 2 and line 3 at busbar 2 (at substation 3)}
    \item \textit{$\mathcal{T}3$: $\mathcal{T}1$ and $\mathcal{T}2$}
    \item \textit{$\mathcal{T}4$: $\mathcal{T}3$ and load 0 and line 4 at busbar 2 (at substation 1)}
    \item \textit{$\mathcal{T}5$: lines 5 and 15 at busbar 2 (at substation 3)}
    \item \textit{$\mathcal{T}6$: $\mathcal{T}1$ and $\mathcal{T}5$}
    \item \textit{$\mathcal{T}7$: lines 6, 15 and 16 at busbar 2 (at substation 3)}
    \item \textit{$\mathcal{T}8$: $\mathcal{T}1$ and $\mathcal{T}7$}
\end{itemize}}
\label{tab:topologies}
\begin{tabular}{|l|l|l|}
\hline
\textbf{\begin{tabular}[c]{@{}l@{}}Number \\ of  episodes\end{tabular}} & \textbf{\begin{tabular}[c]{@{}l@{}}Topology \\ sequence\end{tabular}} & \textbf{\begin{tabular}[c]{@{}l@{}}Entropy \\ range\end{tabular}} \\ \hline

7 & $\mathcal{B}$ & 0 \\ \hline
305 & $\mathcal{B}\rightarrow\mathcal{T}1$ & 0.0043-0.6899\\ \hline
37 & $\mathcal{B}\rightarrow\mathcal{T}2$ & 0.5698-0.6896\\ \hline
431 & $\mathcal{B}\rightarrow\mathcal{T}2\rightarrow\mathcal{T}3$ & 0.0085-1.0878\\ \hline
170 & $\mathcal{B}\rightarrow\mathcal{T}1\rightarrow\mathcal{T}3$ & 0.5734-1.0865\\ \hline
1 & $\mathcal{B}\rightarrow\mathcal{T}1\rightarrow\mathcal{B}\rightarrow\mathcal{T}2$ & 1.0678\\ \hline
1 & $\mathcal{B}\rightarrow\mathcal{T}1\rightarrow\mathcal{T}3\rightarrow\mathcal{T}2\rightarrow\mathcal{T}3$ & 1.235\\ \hline
1 & $\mathcal{B}\rightarrow\mathcal{T}2\rightarrow\mathcal{T}3\rightarrow\mathcal{T}2\rightarrow\mathcal{T}3$ & 0.6767\\ \hline
5 & $\mathcal{B}\rightarrow\mathcal{T}2\rightarrow\mathcal{T}3\rightarrow\mathcal{T}4$ & 1.1696-1.3241\\ \hline
5 & $\mathcal{B}\rightarrow\mathcal{T}1\rightarrow\mathcal{T}3\rightarrow\mathcal{T}4$ & 0.8621-1.0957\\ \hline
1 & $\mathcal{B}\rightarrow\mathcal{T}5\rightarrow\mathcal{T}6\rightarrow\mathcal{T}3$ & 1.0254\\ \hline
1 & $\mathcal{B}\rightarrow\mathcal{T}7\rightarrow\mathcal{T}8$ & 0.365\\ \hline
\end{tabular}
\end{table}

\section{Discussion}
This study presents a simple deep RL approach for power flow control
that is both easy to understand and easy to implement. 
It shows good convergence behaviour, 
it generalises well despite training on only a single well-chosen scenario,
and is self-contained with few tuneable parameters. 
This is in contrast to previous studies where more complicated RL algorithms are employed,
a large amount of scenarios is used for training,
and additional techniques like imitation learning and guided exploration
are needed (e.g \cite{Lan2019,marot2019learning}).
Hence, the RL approach presented here is well-suited as a baseline for future studies on larger and more realistic power networks.

Moreover, we present a first detailed analysis of the control behaviour of the agent
whereas previous studies mainly focused on convergence and performance measures.
We find that the agent generally identifies 2 specific substations out of 7 controllable substations
for optimal power flow control. 
Most scenarios can be successfully operated by applying only 1-2 topological changes
which suggests that the agent acts rather efficiently (i.e. avoids unnecessary activity).
Regarding the timing of the topological changes the simple agent shows
very scenario-dependent behaviour.

In future work, these results  can be expanded upon in different directions, two of which are mentioned below. From a power system perspective, it would be interesting to further analyse the topologies proposed by the agent, to check whether these include improved `base' topologies, or whether switching actions are unavoidable to avoid line overloading in specific scenarios. From a RL perspective, it is interesting to investigate how the generalisation behaviour of the proposed approach can be further improved, without sacrificing the overall simplicity of the method.

\bibliographystyle{IEEEtran}

\end{document}